%% file: main.tex
\documentclass[10pt,twocolumn,letterpaper]{article}

\usepackage{cvpr}              
\input{preamble}  

\definecolor{cvprblue}{rgb}{0.21,0.49,0.74}
\usepackage[pagebackref,breaklinks,colorlinks,allcolors=cvprblue]{hyperref}

\usepackage{multirow}

\def\paperID{9}
\def\confName{EDGE}
\def\confYear{2025}

\title{Scaling On-Device GPU Inference for Large Generative Models}

\author{
  Jiuqiang Tang\thanks{Corresponding authors:~\scriptsize{\tt\{jqtang,sorokin,impjdi\}@google.com}}\\Google LLC \and
  Raman Sarokin\footnotemark[1]\\Google LLC \and
  Ekaterina Ignasheva\thanks{Work done while at Google LLC}\\Meta Platforms, Inc.\and
  Grant Jensen\\Google LLC \and
  Lin Chen\\Google LLC \and
  Juhyun Lee\footnotemark[1]\\Google LLC \and
  Andrei Kulik\\Google LLC \and
  Matthias Grundmann\\Google LLC
}

\begin{document}
\maketitle

\begin{abstract}

Driven by the advancements in generative AI, large machine
learning models have revolutionized domains such as image processing, audio
synthesis, and speech recognition.  While server-based deployments remain
the locus of peak performance, the imperative for on-device inference,
necessitated by privacy and efficiency considerations, persists.
Recognizing GPUs as the on-device ML accelerator with the widest reach,
we present ML Drift--an optimized framework that extends the
capabilities of state-of-the-art GPU-accelerated inference engines.
ML Drift enables on-device execution of generative AI workloads which
contain 10 to 100$\times$ more parameters than existing on-device generative AI
models.
ML Drift addresses intricate engineering challenges associated with cross-GPU API development, and ensures
broad compatibility across mobile and desktop/laptop platforms,
thereby facilitating the deployment of significantly more complex models
on resource-constrained devices.  Our GPU-accelerated ML/AI inference
engine achieves
an order-of-magnitude performance improvement relative to existing
open-source GPU inference engines.

\end{abstract}

\section{Introduction}

The past decade has witnessed a rapid proliferation of large generative models
revolutionizing domains such as image synthesis and natural language processing,
particularly on the server-side.
While server-based
deployments offer substantial computational resources, on-device execution
remains crucial for
user privacy, reduced latency, offline functionality, and lower server costs.
Among popular on-device processing units, such as CPUs, GPUs, and NPUs,
mobile GPUs stand out due to their widespread availability and
intrinsic computational capabilities, thus presenting a ubiquitous
solution for the deployment of these complex models.  However,
modern generative models, distinguished by their substantial scale,
often 10 to 100$\times$ larger in parameters than their predecessors,
present significant engineering challenges for deployment on
resource-constrained mobile GPUs.
Existing on-device inference frameworks struggle to manage the increased
memory and computational demands, which can be one to two orders of
magnitude greater, leading to performance bottlenecks and limited scalability.

This paper introduces ML Drift, an optimized inference framework designed
to address the deployment challenges of large generative models on GPUs.
Key innovations include: (1) 
tensor virtualization, decoupling logical indices from physical GPU indices
and incorporating coordinate translation for flexible memory layout and
kernel optimization;
(2)
comprehensive optimizations for diverse GPUs, including device specialization
via backend-specific shaders, memory management, operator fusion, and
stage-aware LLM optimizations.
Furthermore, this paper presents a thorough
performance evaluation of ML Drift across mobile (Arm and Qualcomm), desktop (Intel and
NVIDIA), and Apple GPUs, showcasing an order-of-magnitude
improvement over existing open-source engines.

\begin{table}[!t]
  \centering
  \begin{tabular}{@{}clrr@{}}
  
  \toprule  
  
  \multicolumn{2}{c}{Large Generative Model} &  
  \multicolumn{1}{c}{Mobile} &
  \multicolumn{1}{c}{Laptop} \\
  
  \midrule  
  
  \multicolumn{2}{c}{\multirow{2}{3cm}{Stable Diffusion 1.4\\\footnotesize{$512\times 512$, 20 it., FP16}}} &
  \multirow{2}{8.5mm}{8.97 s} &
  \multirow{2}{8.5mm}{3.40 s}\\ \\
  
  \midrule  
  
  \multirow{2}{3cm}{\hspace{4.5mm}Gemma2 2B\\\footnotesize{\hspace{4.5mm}mixed-q8/4/4}} & \hspace{-0.8cm} \small{prefill}   & 1370 tokens/s & 3920 tokens/s\\
                                                                                        & \hspace{-0.85cm} \small{decode}      & 37.1 tokens/s & 45.7 tokens/s\\
                                                                                        
  \midrule  
  
  \multirow{2}{3cm}{\hspace{4.5mm}Llama3.1 8B\\\footnotesize{\hspace{4.5mm}mixed-q8/4/4}} & \hspace{-0.8cm} \small{prefill} &  412 tokens/s & 1280 tokens/s\\
                                                                                          & \hspace{-0.85cm} \small{decode}    & 12.7 tokens/s & 22.9 tokens/s\\
  \bottomrule  
  \end{tabular}
  \caption*{
  ML Drift GPU inference performance on mobile (Qualcomm Adreno 750) and laptop (Intel Ultra 7 258V) GPUs.
  }
\end{table}


\section{Related Work}

The resource-constrained nature of edge devices presents significant challenges
to the effective deployment of machine learning models.  This limitation
has spurred considerable research into various strategies aimed at optimizing
inference performance within these environments.  These efforts can be broadly
categorized into approaches leveraging general-purpose GPUs, specialized
inference frameworks, and, more recently, investigations into the complexities
of deploying large generative models on edge devices.

\paragraph{General-Purpose GPU Inference}
Much of the existing work on GPU-accelerated inference relies on vendor-specific libraries like TensorRT~\cite{tensorrt} and ROCm~\cite{rocm}. While effective in maximizing performance on their respective hardware platforms, these solutions inherently suffer from architectural specificity, limiting their portability across diverse GPU ecosystems. ML Drift distinguishes itself by prioritizing performance optimization across a broad spectrum of backends, such as OpenCL~\cite{sochacki2019opencl}, Metal~\cite{apple2014metal}, OpenGL ES~\cite{leech2016opengl}, Vulkan~\cite{vkspec}, and WebGPU~\cite{webgpu},
exceeding the scope of other solutions~\cite{Lee19,Loc2017DeepMonMG}.

\paragraph{Inference Engines for Heterogeneous Edge Devices}
The optimization of inference engines for edge devices, particularly leveraging heterogeneous hardware (CPUs, GPUs, and NPUs), has been a primary research focus. Hardware and operating system vendors have responded to such demand by offering specialized SDKs designed to maximize inference performance on their proprietary hardware. Notable examples of these vendor-specific SDKs include Apple CoreML~\cite{coreml}, Arm Compute Library~\cite{arm-compute-library}, Huawei HiAI Engine~\cite{huawei}, Intel OpenVINO~\cite{openvino}, MediaTek NeuroPilot~\cite{neuropilot}, Microsoft DirectML~\cite{directml}, and Qualcomm Neural Processing SDK~\cite{qairt}. While these SDKs offer significant performance advantages within their designated ecosystems, their inherent vendor dependence raises concerns regarding cross-platform portability and deployment flexibility. In contrast to these vendor-centric approaches, a variety of cross-platform frameworks have emerged, aiming to provide broader hardware and platform coverage. These frameworks strive to abstract away hardware complexities and offer more generalized deployment solutions. Prominent examples of these vendor-agnostic frameworks include ExecuTorch~\cite{executorch}, ONNX Runtime Mobile~\cite{onnxruntime-mobile}, MACE~\cite{xiaomi-mace}, MNN~\cite{ali-mnn}, NCNN~\cite{tencent-ncnn}, and TensorFlow Lite~\cite{tflite}. Additionally, IR-based runtimes, such as IREE~\cite{iree}, TVM~\cite{tvm}, and XLA~\cite{xla}, focus on optimizing machine learning models via lowering them through intermediate representation to hardware-specific code.

\paragraph{Large Generative Model Inference}
The recent emergence of large generative models has further intensified the demands on machine learning inference. In response, the machine learning inference community has proposed specialized libraries tailored to server-side deployment, including LMDeploy~\cite{2023lmdeploy}, SGLang~\cite{zheng2024sglangefficientexecutionstructured}, and vLLM~\cite{kwon2023efficient}. Concurrently, efforts are underway to enable efficient edge inference for large generative models, as evidenced by libraries like llama.cpp~\cite{Llama-cpp}, ollama~\cite{ollama}, and torchchat~\cite{torchchat}. MLC LLM~\cite{mlc-llm} leverages the TVM runtime and WebLLM~\cite{ruan2024webllmhighperformanceinbrowserllm} to facilitate large language model inference across various GPU backends. Specific implementation-level optimizations for large diffusion models on mobile GPUs are also being explored~\cite{chen2023speedneedondeviceacceleration}. Other inference acceleration research ~\cite{alizadeh2024llmflashefficientlarge, Xu2024FastOL, Xue2024PowerInfer2FL, xu2023llmcadfastscalableondevice} focus on challenges orthogonal to our work, including DRAM limitations, NPU/CPU coordination, model collaboration, \etc. Model compression techniques~\cite{lin2023awq, xiao2023smoothquant, frantar2023gptqaccurateposttrainingquantization}, designed to improve edge deployment, are compatible and synergistic with our approach.

\section{Scaling GPU Inference for Large Models}

To address the challenges of deploying large generative models on
diverse GPU architectures, ML Drift extends the basic architecture
of a well-established GPU-accelerated ML inference engine~\cite{Lee19}.
We introduce a novel approach that performs dynamic code generation
at runtime from manually optimized shader templates, categorizing
ML Drift as a specialized inference framework optimizing data layout
and kernel selection.  For easy optimization and performance
exploration, we introduce tensor virtualization, an approach that
abstracts physical GPU objects from logical tensor representations,
enabling diverse memory layouts and kernel configurations, which is
then coupled with coordinate translation, enabling flexible access to
these diverse memory layouts.  Further optimizations include memory
management strategies to reduce footprint, operator fusion,
stage-aware optimizations for LLM inference, and specialized KV
cache layouts.  This architectural paradigm empowers ML Drift to
enhance performance and scalability across a broad spectrum of GPU
platforms.

\subsection{Logical Tensors and Physical GPU Objects}

In the context of this paper, a \textit{logical} tensor refers to a multi-dimensional
array with semantically meaningful axes, as typically conceived in mathematical or
machine learning contexts.  A \textit{physical} GPU object, on the other hand,
is the actual memory buffer or storage structure on the GPU that materializes this
logical tensor,~\eg, GPU buffers, image buffers, texture arrays, 2D textures, and 3D textures.

Since intermediate tensors in a neural network typically do not have inherent
meanings for their axes, we implicitly assign semantics to them.  Specifically,
we assign the following semantics per axis (for tensors up to 5D):
\begin{itemize}
\vspace{-2mm}
\begin{multicols}{2}
    \item \textbf{0D:} Scalar
    \item \textbf{1D:} Linear
    \item \textbf{2D:} $HW$
    \item \textbf{3D:} $HWC$
    \item \textbf{4D:} $BHWC$
    \item \textbf{5D:} $BHWDC$
\end{multicols}
\vspace{-2mm}
\end{itemize}
where $B$, $H$, $W$, $D$, and $C$ represent batch, height, width, depth, and
channel, respectively.
The $D$ axis of 5D tensors is utilized only for 3D convolutions, and for other
networks, $D$ is typically set to $1$.
We have empirically determined the optimal GPU object for each edge device
during offline testing.  At runtime, during initialization, we select the
predetermined optimal GPU object for each GPU kernel based on the detected hardware.
Depending on the chosen GPU object, tensor elements are stored with a
different memory layout and therefore require different indexing methods to
access the corresponding tensor element.

\begin{figure}[!t]
  \centering
  \includegraphics[width=1\linewidth]{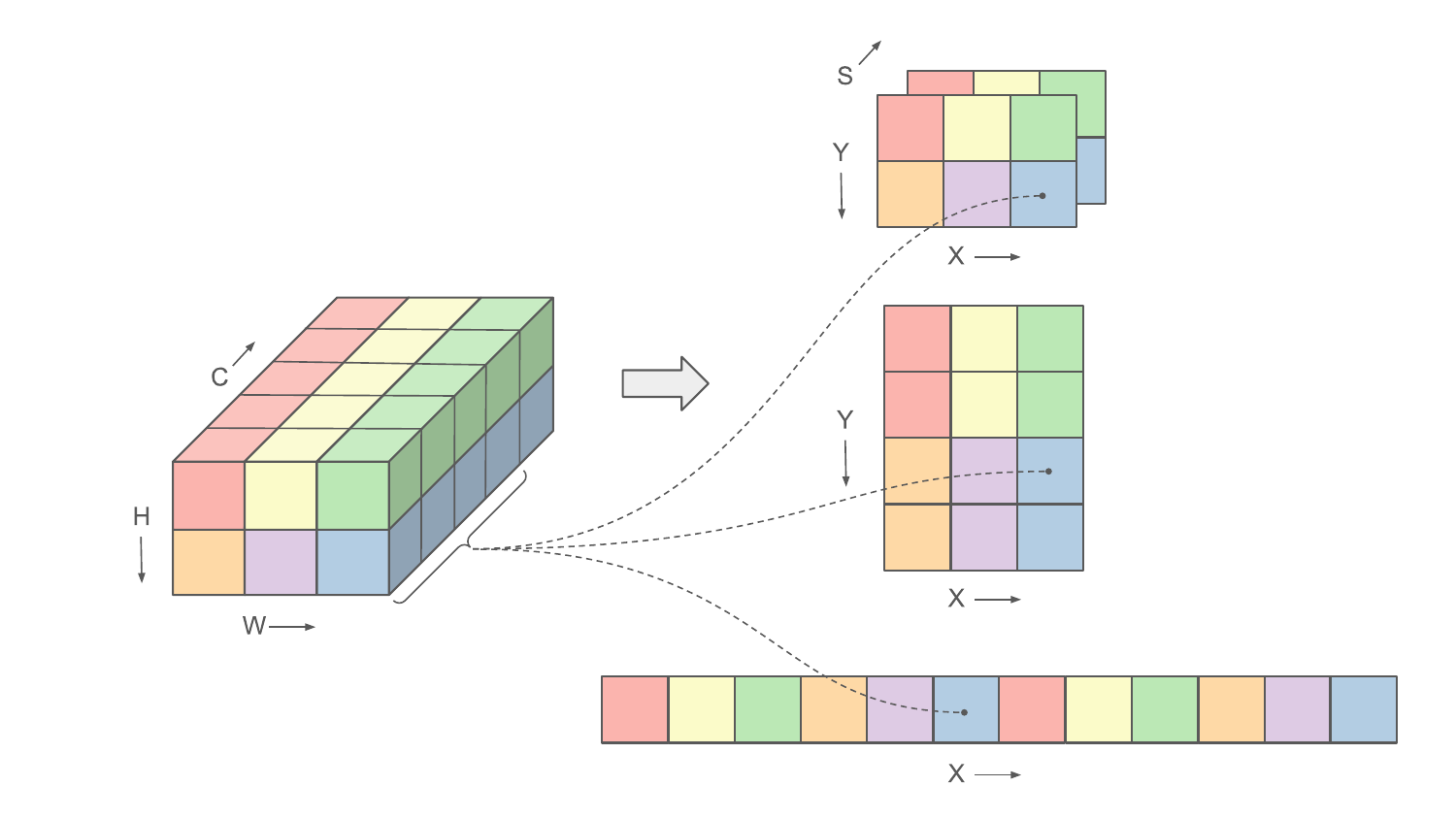}
  \caption{
    Tensor virtualization enables flexible memory layouts. A logical tensor of size
    $(1,2,3,5)$ can be stored as 
       a ``physical'' 3D texture in $DSHWBC_4$ layout (top right), 
       a ``physical'' 2D texture in $HSWBDC_4$ layout (middle right), and
       a ``physical'' 1D image buffer in $DSHWBC_4$ layout (bottom right). The squares on the right side of the illustration represent 4-channel slices.
  }
  \label{fig:memory-layouts}  
\end{figure}

Inspired by the PHWC4 memory layout~\cite{Lee19}, which exploits the GPU's
4-element SIMD by organizing data into contiguous 4-channel slices optimized for
GPU buffers and textures, we employ a diverse set of 4-element slice-aware
memory layouts.
For example, consider a 5D tensor of size $(B,H,W,D,C)$.
When realized with a 2D texture, we use the memory layout $HSWBDC_4$, where $S$
represents slices ($\lceil C/4\rceil$) and $C_4$ represents the index within the slice
($C \bmod 4$), to apply automatic zero clamp for the $H$ dimension.
To ensure compatibility with 4-element SIMD, tensors with channel
count not divisible by 4 are zero-padded.

We perform a similar exploration for the weight tensors for the optimal
GPU object and the memory layout. The most frequently used layouts for weights
of convolutions and fully connected are $OHWI$ or $OHWDI$, where $I$ and $O$ are
the number of input channels and output channels, respectively, which are rearranged 
to a permutation of the $(G, S_O, O_4, HWD, S_I, I_4)$ layout depending on the kernel design. 
Here, $S_I$ represents the number of slices of the
$I$ axis, and $O_4$ and $I_4$ represent the elements inside the slice of $O$ and
$I$, respectively. While the specific values of $G$ and $S_O$ are kernel-design
dependent, their product, $G \cdot S_O$, always corresponds to the number of slices 
of the $O$ axis.

By strategically selecting the optimal memory layout for weight tensors,
we achieve up to a 20\% speedup in matrix multiplication operations, which are
fundamental to both convolution and fully connected layers, compared to using a
naive layout.
This weight tensor layout optimization, fundamental to high-performance matrix
multiplication, is a primary driver of ML Drift's performance advantage.

\subsection{Tensor Virtualization}

Relying on a single uniform layout for all tensors can be suboptimal for certain
GPU kernels whose memory access diverges from the originally intended pattern.
To address this limitation, we introduce ``tensor virtualization,'' a novel
technique inspired by virtual memory.  Tensor virtualization decouples the
logical representation of a tensor from its physical storage on the GPU,
allowing tensors to be realized using various types and numbers of GPU
memory objects (textures, buffers,~\etc.).  An abstraction layer manages the
mapping between logical tensor indices and physical GPU object indices,
handling the underlying fragmentation and distribution of tensor data.  This
frees kernel authors from low-level memory management concerns, enabling them to
focus on algorithm logic.

\begin{figure}[!t]
  \centering
  \includegraphics[width=1\linewidth]{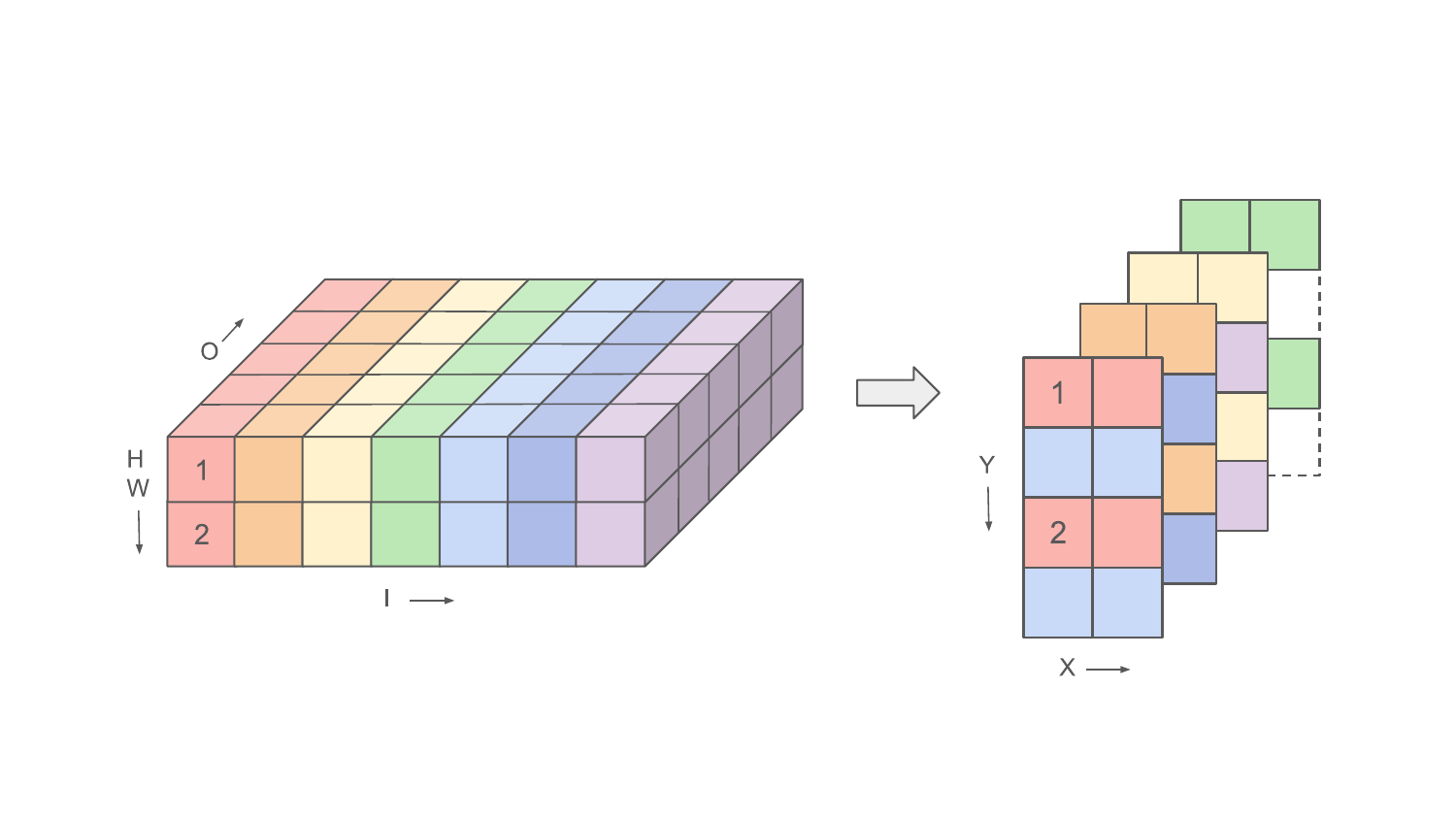}
  \caption{
    Tensor virtualization decouples logical tensor indices from physical GPU
    indices. An $OHWI$ weights tensor of size $(5, 2, 1, 7)$, utilized in a convolution, can be stored as a 2D texture array. This array consists of four 2D textures, each measuring $(4, 2)$, which amounts to 8 vec4 elements per texture. Dotted-line squares indicate padding used for alignment.
  }
  \label{fig:weights-layout}  
\end{figure}

This virtualization provides significant flexibility in tensor representation.
Whereas a tensor with dimensions $(1,2,3,5)$ would traditionally be
rigidly constrained to the PHWC4 format, requiring a RGBA texture of size
$(2,3\times\lceil5/4\rceil)=(2,6)$, our approach allows this logical tensor
to be arbitrarily mapped to various memory layouts. As shown in Figure~\ref{fig:memory-layouts}, it can be
realized as a 3D texture of size $(2, 3, \lceil5/4\rceil)=(2, 3, 2)$, a 2D 
texture of size $(2\times\lceil5/4\rceil,3)=(4,3)$, or as a 1D image buffer with
$2\times3\times\lceil5/4\rceil=12$ pixels, as deemed most efficient by the
specific kernel implementation. This adaptability is crucial for optimizing
performance across diverse kernels with varying parallelism characteristics.

Tensor virtualization also facilitates efficient handling of scenarios
where a single tensor needs to be represented by multiple GPU objects.
For instance, our generic convolution kernel necessitates reading from
multiple textures simultaneously to optimize memory cache usage.
Our abstraction layer seamlessly supports this by allowing a tensor of size
$(5,2,1,7)$ to be represented using four textures of size $(4,2)$, with
each texture holding a slice of the tensor, as illustrated in Figure~\ref{fig:weights-layout}. This enables complex operations
that might not be feasible with a rigid one-to-one mapping between tensors
and GPU objects.

Though tensor virtualization increases implementation complexity, its
performance overhead is negligible.
Specifically, the mapping between logical and physical indices is
established during the shader code generation process at initialization,
thus minimizing runtime latency.

\subsection{Coordinate Translation}

\begin{table}
  \centering
  \begin{tabular}{@{}ll@{}}
    \toprule
    Storage Type & Storage Coordinates \\
    \midrule

    1D Buffer & $((s \cdot \text{height} + y) \cdot \text{width} + x) \cdot \text{batch} + b$ \\
    2D Texture & $(x \cdot \text{batch} + b, y \cdot \text{slice} + s)$ \\
    3D Texture & $(x \cdot \text{batch} + b, y, s)$ \\
    \bottomrule
  \end{tabular}
  \caption{
    Coordinate translation from logical coordinates $(b,x,y,s)$ to GPU
    memory object’s coordinates for a $BHWC$ tensor, considering its
    batch size, width, height, slice count, and storage type.
    \vspace{-3mm}
    }
  \label{tab:coord-translation}  
\end{table}

To enable the aforementioned flexibility for the producing shader program,
the consuming shader programs must be equipped with
the capability to read and write tensors with flexible memory layouts.  
In other words, we introduce a pre-processing stage in the shader code 
gen with helper functions,~\eg,~\texttt{args.src.Read(b,x,y,s)}, 
that translates the requested tensor element's coordinates to the actual
GPU memory object's coordinates underneath, as exemplified in
Table~\ref{tab:coord-translation}.  This enables the shaders
implementing the neural network operators to abstract accessing tensor
elements from GPU buffers and textures, and simplifies the authoring of the shader 
programs without having to worry about the combinatorial explosion of
memory layouts.

Similar to tensor virtualization, the coordinate translation process is also
performed during shader code generation.  This pre-processing approach avoids
adding any runtime latency when the GPU kernels are executed.

\subsection{Device Specialization}
To facilitate execution across diverse GPU architectures, we develop a suite of shader  generators for various GPU backends (\eg, OpenCL, Metal, WebGPU) that transform platform-agnostic abstractions into target GPU languages. Following runtime analysis of the target GPU's properties, specifically, identifying supported vendor extensions and determining the optimal GPU object storage types, the generator performs a series of transformations on each GPU operator. These transformations include:

\begin{itemize}
\setlength{\parskip}{0pt}
\setlength{\itemsep}{0pt plus 1pt}
\item \textbf{Adaptive kernel selection:} Selecting the fastest implementation from a set of candidates for the specific GPU API and device.
Specialized kernels,~\eg, Winograd fast convolutions, may be employed for better performance.
\item \textbf{Exploitation of vendor-specific extensions:} Leveraging vendor extensions such as the specialized matrix multiplication extension,~\eg,~\texttt{cl\_arm\_matrix\_multiply}. 
\item \textbf{Syntax translation:} Converting ML Drift representation into language-specific shader code.
\item \textbf{Weights conversion:} Pre-processing weights and storing them in GPU memory with the optimal memory layout. 
\end{itemize} 

\begin{figure}[!t]
  \centering
  \includegraphics[width=1\linewidth]{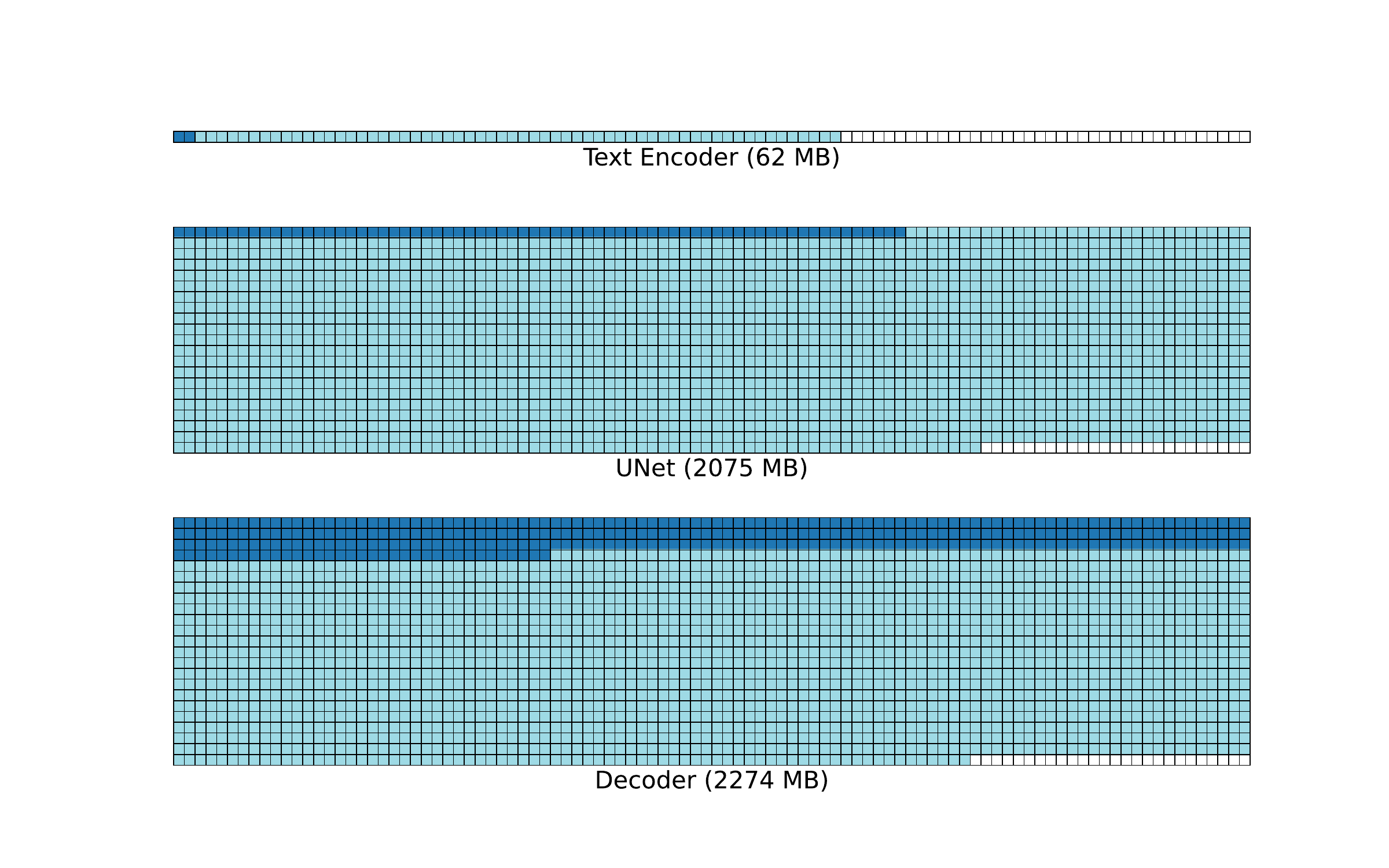}
  \caption{
    Memory savings for Stable Diffusion 1.4 using \textsc{Greedy by Size} policy
    for offset calculation~\cite{pisarchyk2020}.
    Light squares represent naive allocation (text encoder: 62 MB, UNet: 2075 MB, VAE decoder: 2274 MB);
    dark squares represent optimized memory footprint (text encoder: 2 MB, UNet: 65 MB, VAE decoder: 320 MB).
    Each square equals 1 MB,~\ie, each row equals 100 MB.
    \vspace{-3mm}
  }
  \label{fig:mm}  
\end{figure}

\subsection{Memory Management}

The memory footprint of large generative models is substantially influenced by
both model weights and intermediate tensors.  While model weight size serves
as a primary determinant of overall memory consumption, the efficient
management of runtime memory is crucial.  Prior investigations into
intermediate tensor management have demonstrated that, owing to the
sequential execution paradigm of neural networks, these tensors need not
simultaneously occupy memory.  Consequently, the implementation of
efficient memory buffer reuse methodologies presents a significant means
of reducing the runtime memory footprint~\cite{pisarchyk2020}.
Memory sharing can be achieved through two primary approaches:
assigning memory buffers to tensors with non-overlapping lifespans, or
pre-allocating a large memory block and assigning offsets within it to tensors.
These strategies leverage the directed acyclic graph representation and
sequential execution of neural networks.

As an illustrative example, Stable Diffusion 1.4 would require 4.31 GB of runtime
memory for half-precision floating-point (FP16) activations. 
The \textsc{Greedy by Size} strategy reduces the runtime memory
footprint to 387 MB (93\% savings) as shown in Figure~\ref{fig:mm}.

\subsection{Operator Fusion}
\label{sec:op_funsion} 
\begin{figure}[!t]
  \centering
  \includegraphics[width=1\linewidth]{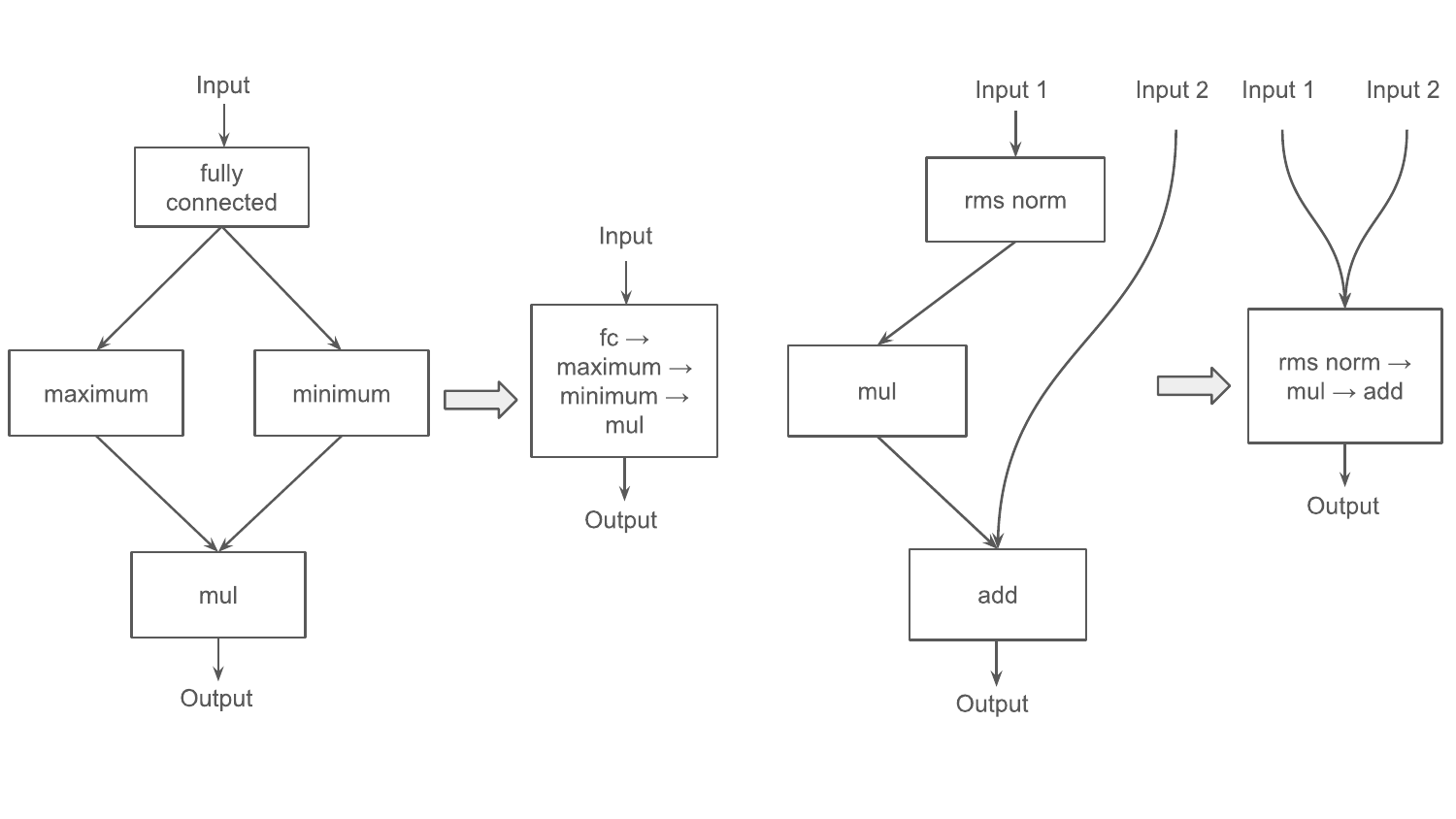}
  \caption{
    Automatic operator fusion examples for large model inference.
    Left: fusion of element-wise operators from two branches with a fully connected operator.
    Right: merging residual connection and element-wise operations with a manually optimized RMS normalization operator.
  }
  \label{fig:op-funsion}  
\end{figure}

Operator fusion, a common optimization, merges multiple memory-bound operations into a single kernel to reduce kernel launch overhead and memory transfers. As shown in Figure~\ref{fig:op-funsion}, ML Drift automatically applies operator fusion when it detects sequences of element-wise operations, tensor reordering operations, or residual connections. Complementing automated optimization strategies, we also incorporate profiling to identify performance bottlenecks and manually implement optimizations for efficient large model inferences. As a specific instance within the attention block, we crafted a custom kernel to combine rotary embedding with the layout transformations of query (Q), key (K), and value (V) projections.
This includes transforming the query projection from an initial layout of
$(B, 1, S, h_q\cdot d_h)$ to a resultant layout of
$(B\cdot h_{kv}, S\cdot h_q/h_{kv}, d_h)$.
Here, $B$ represents batch size,
$h_{kv}$ the number of KV heads,
$S$ sequence length,
$h_q$ the number of query heads,
$d_h$ the head dimension.
This optimized QKV layout is crucial for efficient inference across diverse
attention mechanisms, including multi-head, multi-query, and grouped-query attention.


\subsection{Stage-Aware Optimizations for LLM Inference}
Despite using the same weights residing on the GPU memory, ML Drift distinguishes between the prefill and decode stages of the LLM inference, a differentiation necessitated by their fundamentally disparate performance profiles. For operations involving external weights, such as those calculating the query, key, value, output projections, alongside the linear layers in the feed-forward network, the compute-intensive prefill stage benefits from a dedicated GPU quantization kernel. This kernel converts the floating-point input activations to 8-bit integers and computes the requisite quantization parameters, thereby enabling a subsequent kernel to leverage fast int8 instructions with pre-quantized weights, and to perform dequantization on the output activations. Conversely, the memory-bound decode stage is optimized by integrating input activation quantization directly within the operational kernel, an approach that mitigates memory transfer overhead and consequently enhances aggregate performance. Additionally, ML Drift selects specialized GPU kernels for each stage, contingent upon the characteristics of the input data.
The prefill stage, dealing with longer input sequences, benefits from highly optimized convolution kernels,
whereas the decode stage, characterized by the iterative generation of individual tokens, is more effectively served by fully connected kernels.

\begin{figure}[!t]
  \centering
  \includegraphics[width=1\linewidth]{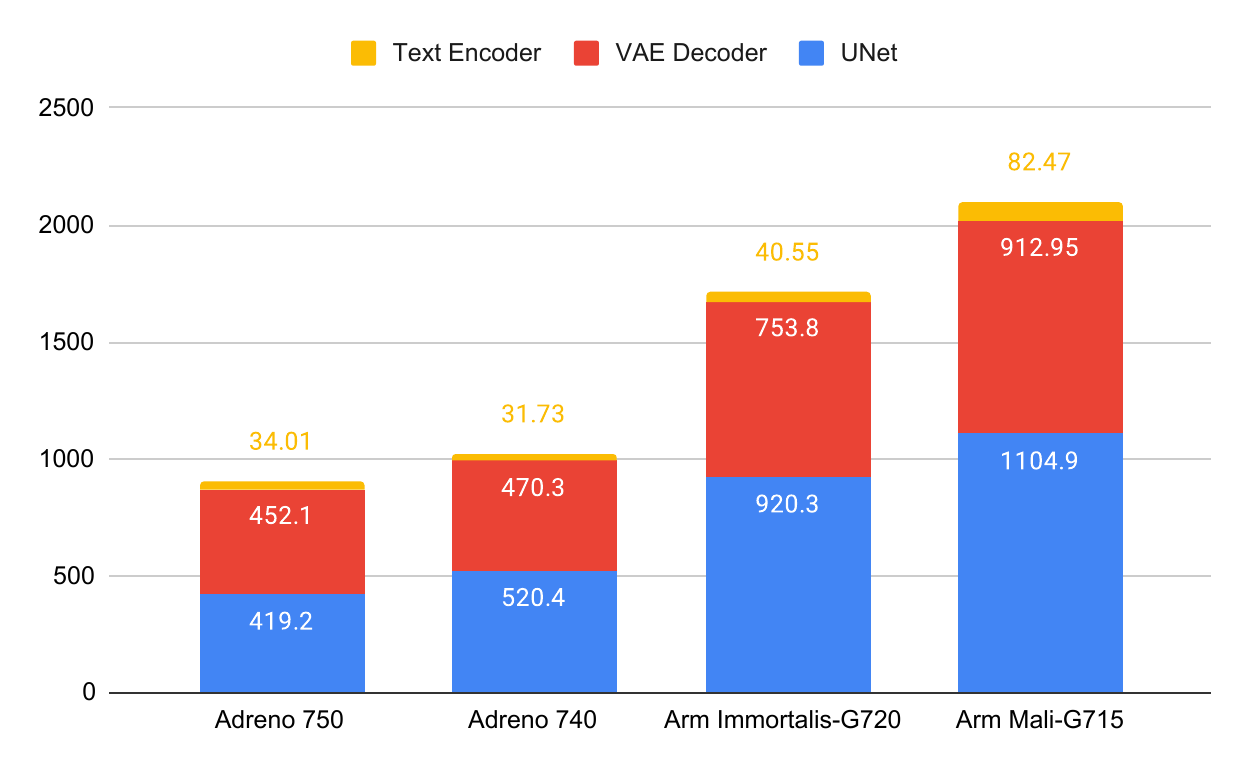}
  \caption{
    Single-step inference latency (milliseconds) for Stable Diffusion 1.4, by
    model component (text encoder, VAE decoder, UNet), on Qualcomm and Arm GPUs.
    }
   \label{fig:sd-android}  
\end{figure}

\subsection{GPU-Optimized KV Cache Layout}
ML Drift uses convolution kernels for matrix multiplications in LLM inference.
KV cache, acting as convolution weights, is stored in layouts compatible with the
QKV layout transformation (Section~\ref{sec:op_funsion}).
K cache uses an $OHWI$ layout ($O$=cache\_size, $I$=$d_h$), representing $K^T$ for $QK^T$ computation.
V cache uses $OHWI$ layout with reversed dimensions ($O$=$d_h$, $I$=cache\_size), ensuring the
convolution involving V yields the desired attention output layout
$(B\cdot h_{kv}, S\cdot h_q/h_{kv}, d_h)$.

\section{Performance Evaluation}

This section presents a comprehensive performance evaluation of ML Drift on
large text-to-image models and large language
models across diverse platforms.  We provide detailed benchmarks for ML Drift's
OpenCL, Metal, and WebGPU backends on a range of hardware including
mobile GPUs (Arm Mali and Qualcomm Adreno), desktop/laptop GPUs (Intel and NVIDIA),
and Apple Silicon.  

\subsection{Large Diffusion Models}

We evaluated the performance of our OpenCL, Metal, and WebGPU backends
with the open-source model Stable Diffusion 1.4, configured for FP16 inference with FP16 weights.
This pipeline
includes a text encoder, UNet, and VAE decoder.  Benchmarks were performed
across a range of GPUs from Apple, Arm, Intel, and Qualcomm.

\paragraph{Mobile GPUs}
ML Drift's OpenCL backend serves as the primary execution engine for Android platforms.
On a Samsung S23 Ultra, powered by the Qualcomm Adreno 740 GPU, we achieved an end-to-end
latency of 10.96 seconds for generating $512\times512$ image (20 iterations).
This represents an 8\% improvement over a previously reported sub-12 second
benchmark~\cite{chen2023speedneedondeviceacceleration} and a 26\% gain over
another measurement of 14.42 seconds~\cite{qualcomm-sd}.  Figure~\ref{fig:sd-android}
details Stable Diffusion 1.4 latency on Android, illustrating performance
across individual model components on various mobile GPUs.  
ML Drift OpenCL reduces the
latency to under 9 seconds on a Samsung S24 with Qualcomm Adreno 750.

\begin{table*}[b]
\centering
\begin{tabular}{ @{}l c c c c c c c c c c @{}} 
\toprule
 & \multicolumn{2}{c}{\small Adreno 830} & \multicolumn{2}{c}{\small Adreno 750} & \multicolumn{2}{c}{\small Adreno 740} & \multicolumn{2}{c}{\small Immortalis-G720} &  \multicolumn{2}{c}{\small Mali-G715} \\
 \cmidrule(lr){2-3} \cmidrule(lr){4-5} \cmidrule(lr){6-7} \cmidrule(lr){8-9} \cmidrule(lr){10-11}
 & {\small prefill} & {\small decode} & {\small prefill} & {\small decode} & {\small prefill} & {\small decode} & {\small prefill} & {\small decode} & {\small prefill} & {\small decode}\\
\midrule 
Gemma 2B q8       & 1440 & 22.8 & 1440 & 23.1 & 1120 & 20.4 & 1280 & 18.2 & 796 & 11.9 \\
Gemma 2B 8/4/4    & 1490 & 42.5 & 1480 & 42.7 & 1150 & 38.1 & 1380 & 32.5 & 813 & 12.2 \\
\midrule  
Gemma2 2B q8      & 1220 & 20.8 & 1290 & 21.3 & 1010 & 18.3 & 1170 & 15.7 & 700 & 11.2 \\
Gemma2 2B 8/4/4   & 1250 &	37.0 & 1370 & 37.1 & 1040 & 32.4 & 1250 & 27.3 & 729 & 18.4 \\
\midrule
Llama3.2 3B q8    &  960 & 17.1 &  917 & 17.5 &  720 & 15.4 &  791 & 12.5 & 507 & 8.71 \\
Llama3.2 3B 8/4/4 & 983 & 30.4 &  959 & 30.3 &  741 & 26.8 &  850 & 21.2 & 516 & 15.0 \\
\midrule
Llama3.1 8B q8	  & 389 & 7.70 &  - & - & - & - & 270	& 4.72 & - & - \\	
Llama3.1 8B 8/4/4 & 413 & 13.4 &   412 & 12.7 & 325 & 10.7 &  378 & 8.88 & 240 & 6.46 \\
\bottomrule
\end{tabular}
\caption{
    LLM performance (tokens/s) on Qualcomm and Arm GPUs for Gemma and Llama models.
    Llama 3.1 8B q8 terminated due to memory limitations on devices with Adreno 750, Adreno 740, and Mali-G715.
}
\label{table:mld-opencl-llm-per-device}  
\end{table*}

\begin{table}[!h]
  \centering
  \begin{tabular}{@{}lccc@{}}
    \toprule
    \multicolumn{1}{c}{} & {\small ML Drift} & {\small ML Drift} & {\small ONNX Runtime} \\
    \multicolumn{1}{c}{} & {\small OpenCL}   & {\small WebGPU}   & {\small DirectML} \\
    \midrule
    per iteration & 0.64 & 1.28 & 1.75 \\
    end-to-end    & 13.5 & 27.9 & 37.0 \\
    \bottomrule
  \end{tabular}
  \caption{Stable Diffusion 1.4 performance (seconds) on Intel Meteor Lake Ultra 7 165U.
  \vspace{-5mm}
  }
  \label{tab:sd-intel-165u}  
\end{table}

\paragraph{Desktop/Laptop GPUs}
On a Windows laptop with the Intel Meteor Lake Ultra 7 165U, our
OpenCL and WebGPU backends demonstrated significant performance advantages over ONNX Runtime with DirectML~\cite{sd-olive-directml}.  Specifically, as detailed
in Table~\ref{tab:sd-intel-165u}, ML Drift OpenCL achieved $2.7\times$
speedup, while WebGPU delivered a $1.3\times$ speedup.

To further highlight ML Drift's capabilities, we conducted a targeted comparison
with Intel's reported performance on the newer Lunar Lake Ultra 7
288V~\cite{intel-core-ultra-media-deck}.  On our closest available platform, the
Lunar Lake Ultra 7 258V, ML Drift OpenCL generated a
$512\times512$ image (20 iterations) in 3.4 seconds, representing a 14.4\% speedup
over Intel's reported 3.89 seconds for the 288V.

\paragraph{Apple Silicon}
ML Drift's Metal backend provides optimized execution on Apple Silicon, resulting 
in faster diffusion model inference. Performance testing with Stable
Diffusion 1.4 on a M1 Ultra and MacBook Pro M4 Pro (20-core GPU) yielded runtimes
of 3.86 seconds and 5.34 seconds, respectively. These figures demonstrate a
considerable performance advantage over 5.03 seconds and 6.16 seconds observed
in Apple's CoreML Stable Diffusion~\cite{apple-coreml-sd}.

\begin{figure}[!t]
  \centering
  \includegraphics[width=1\linewidth]{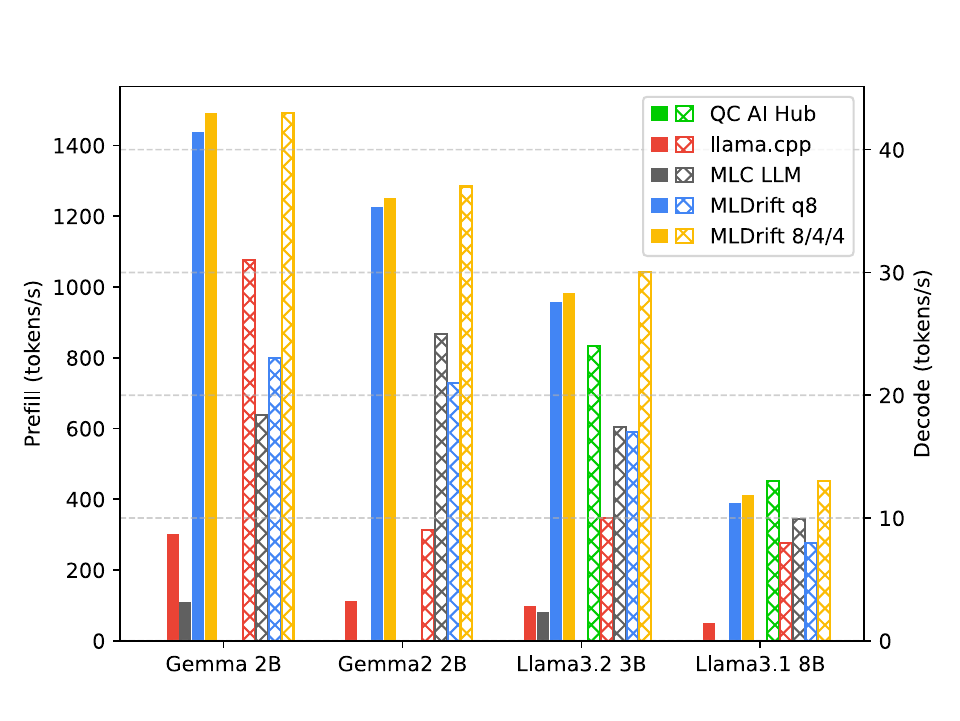}
  \caption{
  Comparative analysis of LLM performance (tokens/s) on Adreno 830,
  evaluating different inference solutions.
  Prefill performance (solid bars, left ordinate) and decode performance
  (cross-hatched bars, right ordinate) are presented.
  Bars are absent when data is unavailable, non-public, or unreasonable.
  }
  \label{fig:ardeno_830_benchmark}  
\end{figure}

\subsection{Large Language Models}

To evaluate ML Drift's inference performance on large language models, we
conducted benchmarks using four open-weight models:
Gemma 2B~\cite{gemmateam2024gemmaopenmodelsbased},
Gemma2 2B~\cite{gemmateam2024gemma2improvingopen},
Llama 3.2 3B~\cite{llama-3-2}, and
Llama 3.1 8B~\cite{llamateam2024llama3herdmodels}.
These assessments spanned mobile and desktop platforms,
enabling comparisons with established LLM GPU inference solutions like
llama.cpp~\cite{Llama-cpp}, ollama~\cite{ollama}, torchchat~\cite{torchchat}, MLC
LLM~\cite{mlc-llm}, and MLX LM~\cite{mlx-lm}. ML Drift primarily utilizes FP16 for activations,
with the exception of NVIDIA platforms, where single-precision floating-point
(FP32) is used due to OpenCL driver limitations. ML Drift implements two
quantization strategies: q8 (per-channel int8 quantization of all weights) and
8/4/4 (mixed-precision per-channel quantization, int8 for attention and int4 for
embedding/feed-forward weights). In contrast, other open-source solutions often 
utilize GGUF~\cite{gguf} q4 group quantization, which produces a model size that 
falls between those resulting from ML Drift’s q8 and 8/4/4 methods. For the LLM
benchmark, the ML Drift implementation turned off
speculative decoding~\cite{leviathan2023fastinferencetransformersspeculative}
and flash attention~\cite{dao2022flashattentionfastmemoryefficientexact}, and performed CPU/GPU synchronization after each token generation.
All evaluations used a fixed context length of 1280 tokens, comprising 1024
prefill and 256 generation tokens.

\paragraph{Mobile GPUs}
We benchmarked ML Drift's OpenCL implementation against llama.cpp's 
benchmark tool and the MLC Chat demo app on Android devices,
evaluating performance across five mobile GPUs:
Qualcomm Adreno 830 (Xiaomi 15 Pro 16GB RAM),
Adreno 750 (Samsung S24 8GB RAM),
Adreno 740 (Samsung S23 Ultra 8GB RAM),
Arm Immortalis-G720 (Vivo X100 Pro 16GB RAM), and
Arm Mali-G715 (Google Pixel 9 12GB RAM).

On Qualcomm Adreno GPUs, as shown in Figure~\ref{fig:ardeno_830_benchmark},
ML Drift's OpenCL backend achieved a 5$\times$ to 11$\times$ speedup in token
prefill compared to the other open-source LLM inference solutions. Furthermore,
ML Drift's OpenCL backend outperformed the Qualcomm AI Hub
benchmark~\cite{qualcomm-Llama3b} by 29\% in token generation speed for
Llama 3.2 3B, utilizing the same 8/4/4 quantization scheme.
Similarly, on Arm Mali GPUs, where llama.cpp is currently not  supported, ML 
Drift demonstrated comparable speedups in both prefill and generation stages 
compared to MLC LLM. To illustrate, when running Llama3.2 3B,
ML Drift reached 791 tokens/s in prefill and 12.5 tokens/s in 
decode on average with q8 quantization, while MLC LLM achieved 89.2 and 11.2 
tokens/s with q4f16, respectively.

\begin{figure}[!t]
  \centering
  \includegraphics[width=1\linewidth]{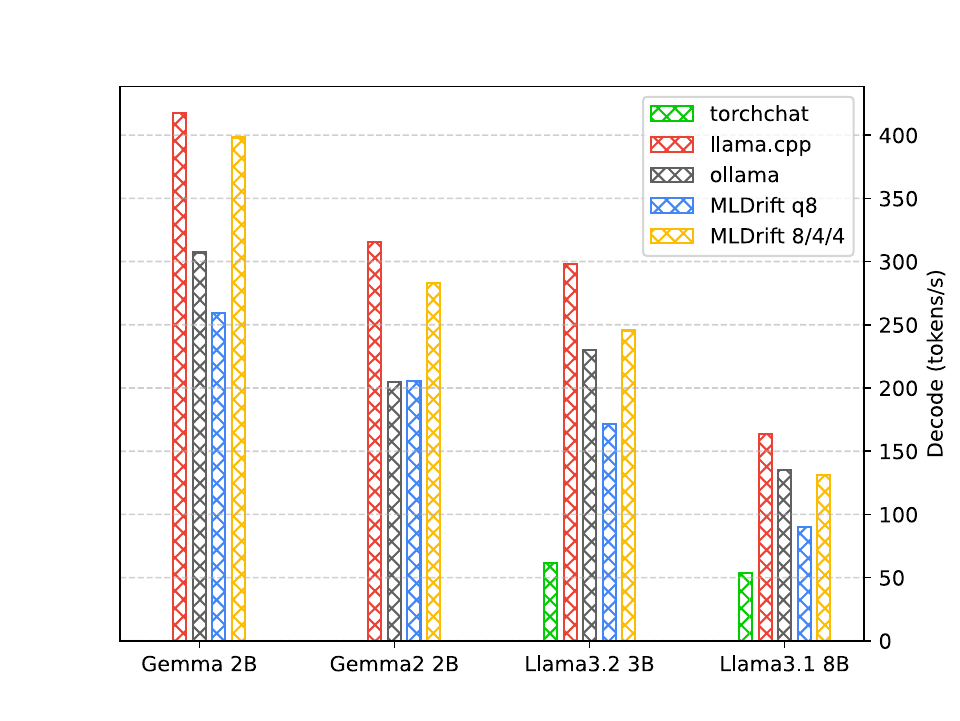}
  \caption{
  Comparative analysis of LLM decode performance (tokens/s) on NVIDIA GeForce RTX 4090,
  evaluating different inference solutions.  Prefill performance is excluded as NVIDIA
  Tensor Cores, which dominate the prefill phase in CUDA, are inaccessible through
  OpenCL APIs, preventing a meaningful comparison.
  }
  \label{fig:rtx4090_decoding}  
\end{figure}

Token prefill speed was largely unaffected by the quantization schema, implying a
compute-bound process.  However, token generation speed demonstrated up to a
1.9$\times$ performance gain with quantization optimization, suggesting a
memory-bound stage where memory bandwidth is a critical constraint.
Comprehensive performance metrics across a range of GPUs and models
are provided in Table~\ref{table:mld-opencl-llm-per-device}.

\paragraph{Desktop/Laptop GPUs}

Extending beyond mobile platforms, ML Drift facilitates execution on
both integrated and dedicated GPUs through OpenCL and WebGPU (native
execution through Dawn~\cite{dawn}) interfaces. Despite the
presence of specialized matrix multiplication acceleration units
inherent in these GPUs, such as NVIDIA's Tensor Cores~\cite{tensor-core},
the current ML Drift implementation is constrained in its ability
to exploit these hardware features, a limitation stemming from
restricted vendor support for these functionalities within OpenCL
and WebGPU.  Consequently, the compute-bound token prefill stage
experiences a fourfold to sevenfold performance decrement, as empirically
observed.  Notwithstanding this limitation, ML Drift sustains
competitive token generation speeds, evidencing its robustness
even in the absence of hardware acceleration.  Specifically, as
illustrated in Figure~\ref{fig:rtx4090_decoding}, on an NVIDIA GeForce
RTX 4090, ML Drift's OpenCL implementation (employing FP32 precision) 
exhibits a 5\% to 25\% performance reduction compared to CUDA-backed llama.cpp,
as measured by llama.cpp's benchmark tool. However, ML
Drift's OpenCL maintains a performance advantage over ollama
and torchchat when performing inference with CUDA backends for
q4f16 quantized models. For the GPUs that support 8-bit cooperative matrix extensions,
\eg, the Intel Ultra 7 platforms, ML Drift exhibits faster token 
prefill speeds, as shown in Table~\ref{table:mld-opencl-llm-intel}.

\begin{table}[!t]
\centering
\begin{tabular}{ @{}l c c c c c c c c c c @{}} 
\toprule
 & \multicolumn{2}{c}{\small Ultra 7 165U} & \multicolumn{2}{c}{\small Ultra 7 258V} \\
 \cmidrule(lr){2-3} \cmidrule(lr){4-5}
 & {\small prefill} & {\small decode} & {\small prefill} & {\small decode} \\
\midrule 
Gemma 2B q8       & 412 & 18.8 & 4110 & 37.2  \\
Gemma 2B 8/4/4    & 435 & 32.2 & 4320 & 57.8  \\
\midrule  
Gemma2 2B q8      & 451 & 15.3 & 3760 & 30.9 \\
Gemma2 2B 8/4/4   & 467 & 25.2 & 3920 & 45.7 \\
\midrule
Llama3.2 3B q8    & 302 & 13.7 & 2650 & 27.7  \\
Llama3.2 3B 8/4/4 & 310 & 22.4 & 2750 & 40.8 \\
\midrule
Llama3.1 8B q8	  & 114 & 7.22 & 1080 & 12.3 \\	
Llama3.1 8B 8/4/4 & 120 & 12.5 & 1280 & 22.9 \\
\bottomrule
\end{tabular}
\caption{\label{table:mld-opencl-llm-intel}%
    LLM performance (tokens/s) on Intel Ultra 7 platforms.
}
\end{table}

ML Drift's WebGPU backend provides operational flexibility
through its support of both FP16 and FP32 precision on the
RTX 4090. A comparative assessment demonstrates a discernible
performance decrement in the WebGPU backend relative to ML
Drift's OpenCL implementation during model inference.
Further research is required to investigate the underlying
causes of this performance reduction and to explore potential
optimization strategies.

\paragraph{Apple Silicon}

\begin{figure}[!t]
  \centering
  \includegraphics[width=1\linewidth]{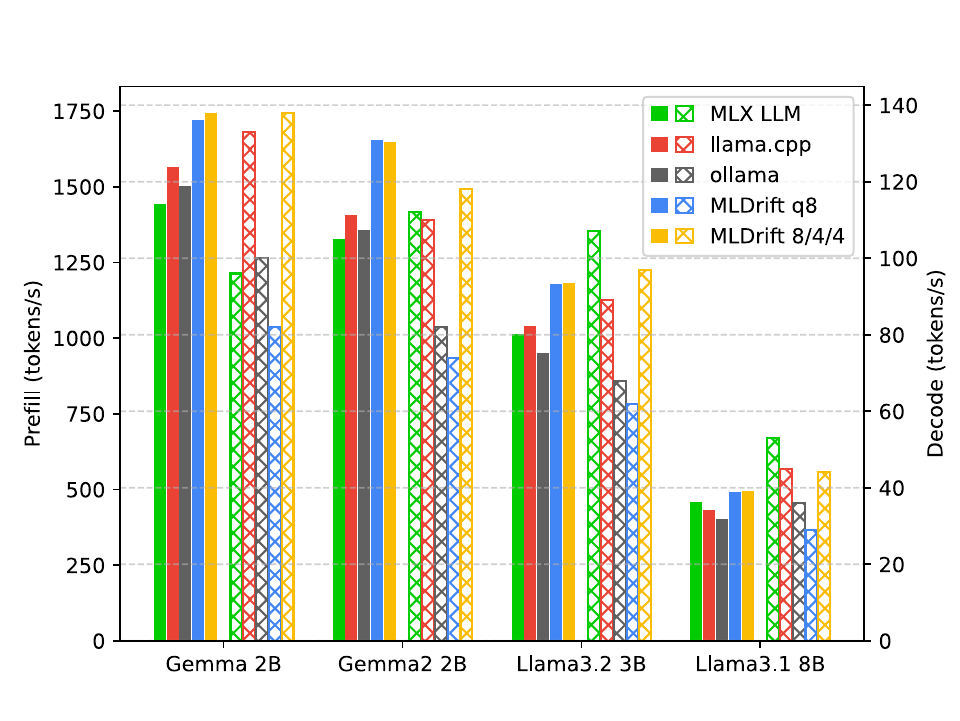}
  \caption{
  Comparative analysis of LLM performance (tokens/s) on Apple M4 Pro (20-core GPU), evaluating different inference solutions.
  Prefill performance (solid bars, left ordinate) and decode performance
  (cross-hatched bars, right ordinate) are presented.
  \vspace{-3mm}
  }
   \label{fig:m4pro_speed}
\end{figure}

For Apple devices equipped with proprietary Apple Silicon
chips, ML Drift facilitates LLM inference via its Metal backend,
typically employing FP16 precision.  The performance of ML
Drift on the Apple M4 Pro 20-core GPU was subjected to rigorous
analysis, with detailed results delineated in Figure~\ref{fig:m4pro_speed}.
The obtained findings
reveal that ML Drift's Metal implementation exhibits a performance
advantage during the token prefill phase. Notably, for Gemma2 2B, ML Drift 
demonstrated speed improvements of 14\% over the llama.cpp benchmark and 20\% 
over MLX LLM. ML Drift consistently outperformed llama.cpp and ollama in token 
generation across all tested models and was also faster than MLX LLM for Gemma 
models.

Consistent with empirical observations on mobile,
a performance disparity in the prefill
phase was discerned when comparing q8 and 8/4/4 quantized
models. While a performance differential between quantization
methodologies persists, its magnitude is attenuated on Apple M4 Pro when contrasted with mobile platforms.
The elevated memory bandwidth characteristic of Apple Silicon's
architecture contributes to the mitigation of the performance variance.

\section{Conclusion}

In this work, we presented ML Drift, a novel inference framework engineered
to facilitate the efficient deployment of large generative models on a
wide range of GPUs.  By decoupling logical tensor indices from physical
GPU indices through tensor virtualization, ML Drift achieves unparalleled
flexibility in memory layout and kernel optimization.  Coupled with device
specialization, memory management strategies, and operator fusion, our
framework delivers substantial performance gains.  A comprehensive
evaluation across diverse hardware platforms, including mobile, desktop,
and Apple Silicon GPUs, validated ML Drift's effectiveness, showcasing
an order-of-magnitude improvement over existing open-source
solutions.  ML Drift demonstrates the capability to execute
workloads one to two orders of magnitude larger than the latest
state-of-the-art GPU inference engines.

Future work will focus on expanding ML Drift's capabilities by
incorporating advanced quantization techniques,~\eg, sub-channel
quantization, and sparsity.  As mobile GPUs increasingly integrate
specialized instructions for ML workloads, including dot products
and matrix multiplications accessible through vendor extensions,
in-depth exploration of these features is crucial for further
performance enhancements.
To facilitate the optimization of ML Drift for diverse models,
an ablation study to quantify the overhead and individual contributions
of each optimization component will be conducted.
Furthermore, evaluations will be extended to more recent diffusion models like Stable Diffusion 3.5~\cite{sd35} and
transformer-based architectures~\cite{vit}, which will involve investigating the
integration of advanced state-of-the-art building blocks tailored for
these model types.
Efficient interoperability and mix-and-match operations with heterogeneous
processors, leveraging zero-copy buffers, 
will also be explored.

{
    \small
    \bibliographystyle{ieeenat_fullname}
    \bibliography{main}
}

\end{document}

%% file: preamble.tex
\usepackage{multicol}
\usepackage{multirow}
\usepackage{stfloats}  
\usepackage{svg}       
